\documentclass[a4paper,twoside]{article}

\usepackage{epsfig}
\usepackage{subcaption}
\usepackage{calc}
\usepackage{amssymb}
\usepackage{amstext}
\usepackage{amsmath}
\usepackage{amsthm}
\usepackage{multicol}
\usepackage{pslatex}
\usepackage{apalike}
\usepackage{algorithm2e}
\usepackage[bottom]{footmisc}
\usepackage{SCITEPRESS}     % Please add other packages that you may need BEFORE the SCITEPRESS.sty package.

\begin{document}

% \title{An Empirical Investigation into Data Quality Aware Approaches for Addressing  Model Drift of Semantic Segmentation Models}

\title{Data Quality Aware Approaches for Addressing  Model Drift of Semantic Segmentation Models}
% \author{\authorname{Samiha Mirza\sup{1}\orcidAuthor{0000-0003-3754-6894}, Vuong D. Nguyen\sup{1}\orcidAuthor{0000-0002-2369-8793}, Shishir K. Shah\sup{1}\orcidAuthor{0000-0003-4093-6906} and Pranav Mantini\sup{1}\orcidAuthor{0000-0000-0000-0000}}
% \affiliation{\sup{1}Quantitative Imaging Lab, University of Houston, Texas, USA}
% \email{smirza6@uh.edu, dnguy222@cougarnet.uh.edu, sshah@central.uh.edu, pmantini@cs.uh.edu}
% }

\title{Data Quality Aware Approaches for Addressing  Model Drift of Semantic Segmentation Models}
\author{\authorname{Samiha Mirza \hspace{1cm} Vuong D. Nguyen \hspace{1cm} Pranav Mantini \hspace{1cm} Shishir K. Shah}
\affiliation{Quantitative Imaging Lab, Dept. of Computer Science, University of Houston, Texas, USA}
\email{smirza6@uh.edu, dnguy222@cougarnet.uh.edu, pmantini@cs.uh.edu, sshah@central.uh.edu}
}
% \author{\authorname{Samiha Mirza\orcidAuthor{0000-0003-3754-6894}, Vuong D. Nguyen\orcidAuthor{0000-0002-2369-8793}, Pranav Mantini\orcidAuthor{0000-0001-8871-9068}, and Shishir K. Shah\orcidAuthor{0000-0003-4093-6906}}
% \affiliation{Quantitative Imaging Lab, Dept. of Computer Science, University of Houston, Texas, USA}
% \email{smirza6@uh.edu, dnguy222@cougarnet.uh.edu, pmantini@cs.uh.edu, sshah@central.uh.edu}
% }

\abstract{ In the midst of the rapid integration of artificial intelligence (AI) into real world applications, one pressing challenge we confront is the phenomenon of model drift, wherein the performance of AI models gradually degrades over time, compromising their effectiveness in real-world, dynamic environments. Once identified, we need techniques for handling this drift to preserve the model performance and prevent further degradation. This study investigates two prominent quality aware strategies to combat model drift: data quality assessment and data conditioning based on prior model knowledge. The former leverages image quality assessment metrics to meticulously select high-quality training data, improving the model robustness, while the latter makes use of learned feature vectors from existing models to guide the selection of future data, aligning it with the model's prior knowledge. Through comprehensive experimentation, this research aims to shed light on the efficacy of these approaches in enhancing the performance and reliability of semantic segmentation models, thereby contributing to the advancement of computer vision capabilities in real-world scenarios.}

\onecolumn \maketitle \normalsize \setcounter{footnote}{0} \vfill

\section{\uppercase{Introduction}}
\label{sec:introduction}

In recent years, the integration of artificial intelligence (AI) into various domains has experienced a remarkable surge, transforming the way we interact with technology. The proliferation of AI applications, ranging from natural language processing to computer vision \cite{khaldi2022unsupervised,Nguyen_2024_CVSL,khan2022deep}, has led to breakthroughs in numerous industries. Among the many AI applications, semantic segmentation plays a pivotal role in object recognition, scene understanding, and image analysis. It aims to assign class labels to individual pixels within an image, thus enabling a deeper understanding of the visual content. Semantic segmentation finds practical application in diverse fields, including autonomous driving, medical imaging, robotics, and remote sensing, among others.

\begin{figure}
\begin{centering}
\includegraphics[width=0.85\columnwidth]{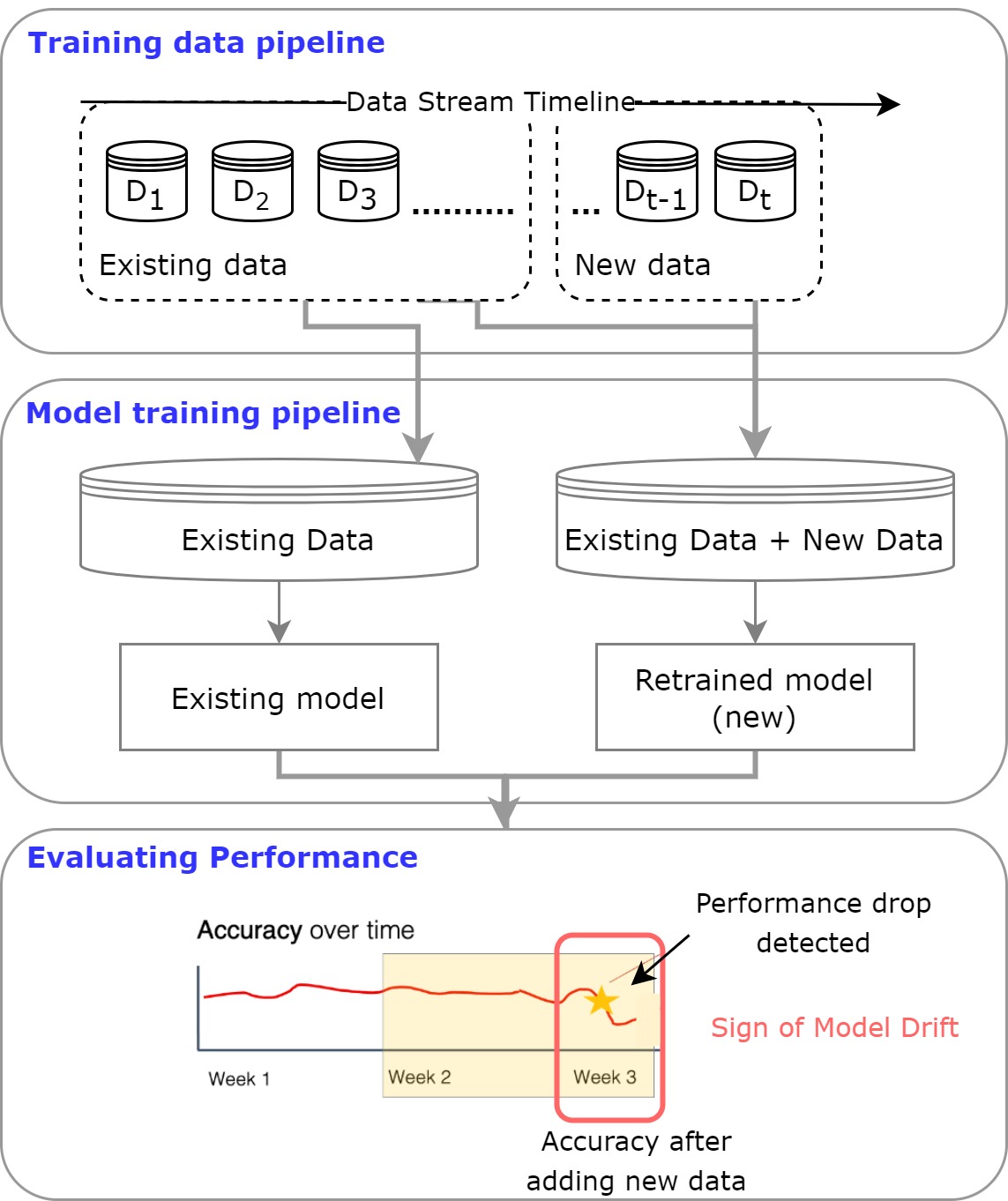}
\par\end{centering}
\caption{Occurrence of model drift. It may happen due to a variety of reasons, with one major reason being the degradation of the quality of the new data.
}
\label{model_dirft}
\end{figure}

\begin{figure*}
\begin{centering}
\includegraphics[width=5.75in, height=2.85in]{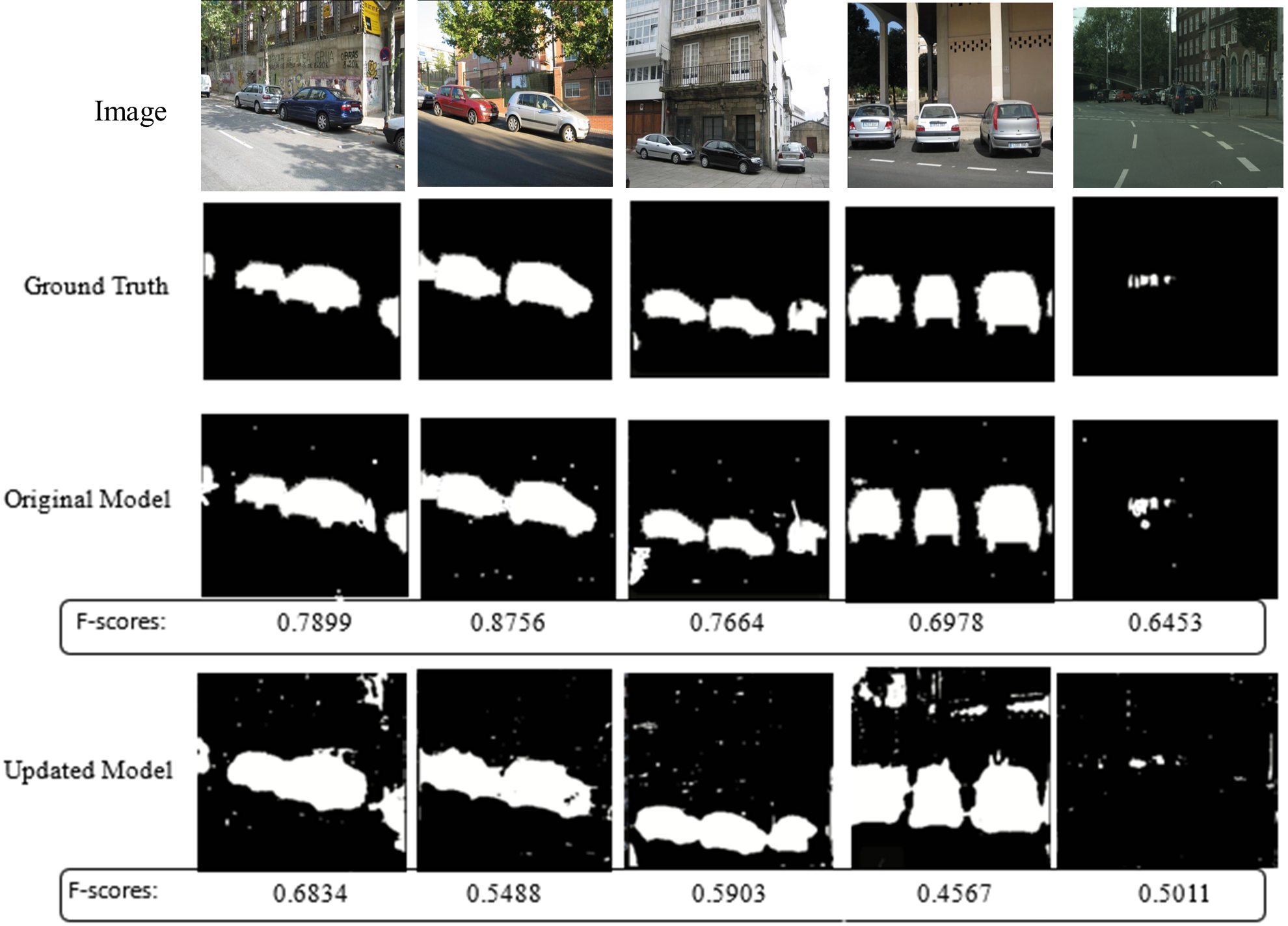}
\par\end{centering}
\caption{Comparison of model predictions between the original model, initially trained with initial datasets, and the updated model, which incorporates both the initial datasets and newly acquired data. The updated model exhibits an increased occurrence of false positives and a corresponding reduction in F-score values when compared to the original model.}
\label{model_degradation}
\end{figure*}

When these segmentation models are trained and deployed in real-world settings, they undergo constant updates and retraining using recently collected datasets in conjunction with their existing training datasets \cite{tsymbal2004problem}. Often, an issue arises when these models are retrained, where their performance tends to deteriorate with the addition of new data, a phenomenon commonly referred to as model drift \cite{whang2023data}. For instance, consider a salt layer segmentation model in the oil and gas industry \cite{devarakota2022deep}. These models are initially trained and deployed for real-time prediction, and are continuously updated (retrained) periodically as more data is gathered. If the new data has a different distribution compared to the original data or is of poor quality, the model performance progressively degrades, resulting in numerous false predictions. This phenomenon, as depicted in Figure \ref{model_dirft}, is a manifestation of model drift, wherein the effectiveness of AI models diminishes over time due to the infusion of new data into the training pipeline \cite{whang2023data}.

Figure \ref{model_degradation} shows an example of model drift in a vehicle segmentation model trained on autonomous driving datasets \cite{zhou2017scene} \cite{zhou2019semantic}, \cite{cordts2016cityscapes}, \cite{Everingham10}. In this case, two models are trained: the first using the original dataset and the latter using both the original and new data. As illustrated in Figure \ref{model_degradation}, we observe a decline in model performance, characterized by an increase in false positives and a subsequent decrease in F-scores following the addition of new data. This prompts the fundamental question of how to select future data that does not compromise model performance.

One reason for the degradation is due to addition of poor quality data. When low-quality or noisy data is included, the model could learn from incorrect or misleading information, leading to suboptimal performance \cite{zha2023data,Nguyen_2024_building}. Another cause for decline in performance is when the distribution of the new data differs from that of the old data and fails to adequately represent the underlying domain \cite{bayram2022concept}. Recognizing these drifts is essential for continuous model maintenance and, once these drifts are recognized, we need techniques to combat and handle this shift to prevent further model degradation.

In this paper, we investigate two major approaches to address the issue of model drift over time. If the model encounters unexpected data quality issues in retraining that were not present in the old training data, it may perform poorly. Thus, spanning from the idea that the performance shift is due to the addition of noisy or distorted data, the first investigated approach considers the intrinsic quality of the data itself \cite{jakubik2022data}. 
% To replicate real world noisy dataset, we first add distortions to the images and employ image quality assessment (IQA) metrics as a criterion for data selection. Investigations on this method highlights the benefits of employing IQA metrics for selecting future data by preserving the model's overall robustness, ensuring that the data used for training aligns with consistent quality standards. Further, our 
Our second approach involves conditioning data selection based on the model's existing knowledge. We use the learned feature vectors from these models as a guide for selecting future data that aligns with its prior knowledge. This creates a more harmonious connection between the model's insights and the selected data.

Our main contributions are as follows: (1) to consider the data quality for updating the models, we propose the use of IQA metrics to select new data for retraining the models; (2) we propose retaining the knowledge from the current production model by selecting future data based on the features learnt by this current model; and (3) we present extensive experiments on multiple benchmark datasets, to highlight the effectiveness of these approaches.

\section{\uppercase{related works}}

% rewrite about types of model drift: concept drift, data drfit, and conclude by mentioning that we are investigating quality aware approaches to model drift
In literature, studies have explored two main types of model drift: concept drift and data drift \cite{bayram2022concept}. Concept drift arises when the statistical properties of the target variable, data distribution, or underlying relationships between variables change over time, rendering previously learned concepts less relevant or outdated \cite{webb2016characterizing}. An example is observed in email spam detection, where evolving spammer tactics can impact model performance over time \cite{guerra2022android}. Data drift, on the other hand, occurs when the statistical properties of new data have changed \cite{ackerman2020detection}. This change may result from significant differences between test and training data or variations in the quality of new data compared to existing training data, which is investigated in this paper.

\subsection{Concept Drift}
Significant research has been conducted on identifying concept drift \cite{lu2018learning}. Wang et al. \cite{wang2022noise} introduced the Noise Tolerant Drift Detection Method (NTDDM) to identify concept drifts in data streams, addressing noise commonly present in real-world applications like those from the Internet of Things. NTDDM employs a two-step approach to distinguish real drifts from noise-induced false alarms. Lacson et al. \cite{lacson2022machine} addressed model drift in a machine learning model for predicting diagnostic testing, employing two approaches: retraining the original model with augmented recent data and training new models. Their findings indicate that training models with augmented data provided better recall and comparable precision. Other researches, such as \cite{wang2015concept}, \cite{dries2009adaptive}, and \cite{klinkenberg2000detecting}, also explore methods to identify concept drift.

\begin{figure*}
    \centering
    \includegraphics[width=6.0in]{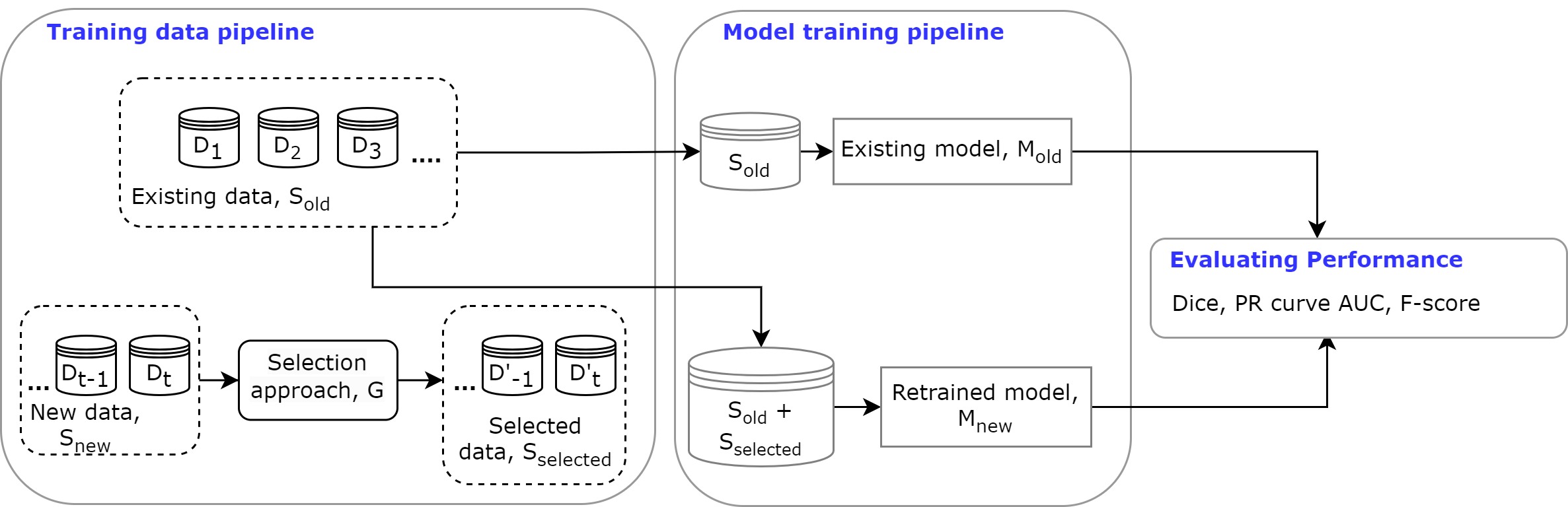}
    \caption{Incorporating quality-aware data selection into the ML pipeline: Existing model $M_{old}$ is trained with $S_{old}$, and new data $S_{new}$ needs integration. Criterion $\mathcal{G}$ explores data selection from $S_{new}$ without causing negative effects, preventing model drift. Two criteria are investigated: data quality-based and feature learning-based. We compare performance between $M_{old}$ and updated model $M_{new}$.}
    \label{Exp_methodology}
\end{figure*}

% Further, Lacson et al. \cite{lacson2022machine} aimed to address model drift in a machine learning model for predicting diagnostic testing, made by the radiologist to the patient, by deploying two approaches: retraining the original model with augmented recent data and training new models. The goal was to assess their performance against the validated original model. They found that training models with augmented data provided better recall and comparable precision. Davis et al. \cite{davis2019nonparametric} employed non parametric based methods to address model drift in clinical prediction setting. For updating the models, their method involved a 2-stage bootstrapping approach where the first stage reduces the overfitting impact on recommendations by generating out-of-bag updated predictions. The second stage assesses these predictions on samples of the same size as the updating set, considering uncertainty linked to the updating sample size in the decision-making process. 

\subsection{Data Drift}

Rahmani et al. \cite{rahmani2023assessing} explored various scenarios of data drift in clinical sepsis prediction, encompassing changes in predictor variable distribution, statistical relationships, and major healthcare events like the COVID-19 pandemic. The study suggests that properly retrained models, particularly eXtreme Gradient Boosting (XGB), outperform baseline models in most scenarios, highlighting the presence of data drift. Davis et al. \cite{davis2019nonparametric} addressed model drift in clinical prediction using non-parametric methods. Their approach involves a two-stage bootstrapping method to update models, mitigating overfitting impact in recommendations. The second stage assesses predictions on samples of the same size as the updating set, considering uncertainty linked to the updating sample size in the decision-making process. Other works, such as \cite{ackerman2021automatically} and \cite{hofer2013drift}, also investigate data drift.

In this paper, we investigate drift due to data quality changes. While prior research has predominantly centered around concept drift, data drift's impact, particularly in segmentation, has been overlooked. We aim to fill this gap by emphasizing the importance of data quality in model pipeline. Notably, our approach uses quality-aware metrics, providing solutions to tackle the challenges associated with data drift and enhancing the robustness of segmentation.

% Most of the research work has investigated techniques to identify and combat concept drift. Limited research has explored data drift, and to the best of our knowledge our paper is the first to incorporate quality aware metrics to handle the model drift in machine learning pipeline more particularly for segmentation models.

% start out by saying that these are the results 

\section{\uppercase{proposed methods}}

\paragraph{Problem formulation:} Our objective is to develop methods for selecting suitable images or data to add to our dataset pipeline for model retraining, all while mitigating degradation in model performance. As illustrated in Figure \ref{Exp_methodology} let's consider that we possess an initial training dataset, denoted as $S_{old} = \{D_1, D_2, \cdots D_n\}$, which is used for training model $M_{old}$. We have new data $S_{new} = \{D_1, \cdots D_m\}$ to incorporate into our pipeline, so we can update our model $M_{new}$, using $S_{old}$ and appropriate data from $S_{new}$. To establish a criterion for selecting the appropriate data $S_{select}$, we need to find a function $\mathcal{G}$ that obtains $S_{select}$ (which prevents model degradation) from $S_{new}$ as follows:
\begin{equation}
S_{selected} = \mathcal{G}(S_{new}),
\end{equation}
where $S_{selected} \subseteq S_{new}$, and $\mathcal{G}$ denotes an approach to select data from $S_{new}$. The following two approaches give our proposed criteria for this selection.

\subsection{Data quality based approach}

This section outlines the methodology for enhancing the performance of a baseline U-Net \cite{ronneberger2015u} model ($M_{\text{old}}$) by incorporating a data quality-based approach. The core idea behind this approach is to leverage quantitative image quality assessment (IQA) metrics to assess the quality of each image within the training dataset and subsequently select a subset of matched quality data from the new data for model refinement. 
% We start by training the baseline U-Net model ($M_{\text{old}}$) using the complete set of available datasets, denoted as $D_{\text{all}}$. The model is trained to optimize a given segmentation task that works on minimizing the cross-entropy loss.
To evaluate the quality of individual images within a set of available datasets $D_{\text{all}}$, we employ IQA metrics. For each image $I_i$ in $D_{\text{all}}$, we compute a quantitative quality value, denoted as $\sigma_i$, using these metrics, computed as:

\begin{equation}
\sigma_i = \mathcal{F}(I_i),
\end{equation}
where, $\mathcal{F}$ represents a function that encapsulates the investigated IQA metrics, generating a scalar value $\sigma_i$ for each image $I_i \in D_{\text{all}}$.

With the set of quality values $ S_{\sigma} =\{\sigma_0 , \sigma_1 , \ldots \sigma_n\}, n = |D_{\text{all}}|$ computed for all images in $D_{\text{all}}$, we proceed to analyze the distribution of these values. We aim to distinguish between high-quality and low-quality images based on a predefined quality threshold $T$. Formally, we define a binary selection function $S(\sigma_i)$ as follows:

\begin{equation}
S(\sigma_i) = 
\begin{cases} 
1, & \text{if } \sigma_i \geq T \\
0, & \text{if } \sigma_i < T 
\end{cases}
\end{equation}

Images for which $S(\sigma_i) = 1$ are considered high-quality and are chosen for model training, while those for which $S(\sigma_i) = 0$ are deemed low-quality and are discarded. This technique relies on an optimal selection of $T$ which avoids two scenarios: (1) too high threshold, which causes the model to stop learning; and (2) too low threshold, which causes the model to capture noise.

\begin{figure*}[h]
    \centering
    \includegraphics[width=6.0in]{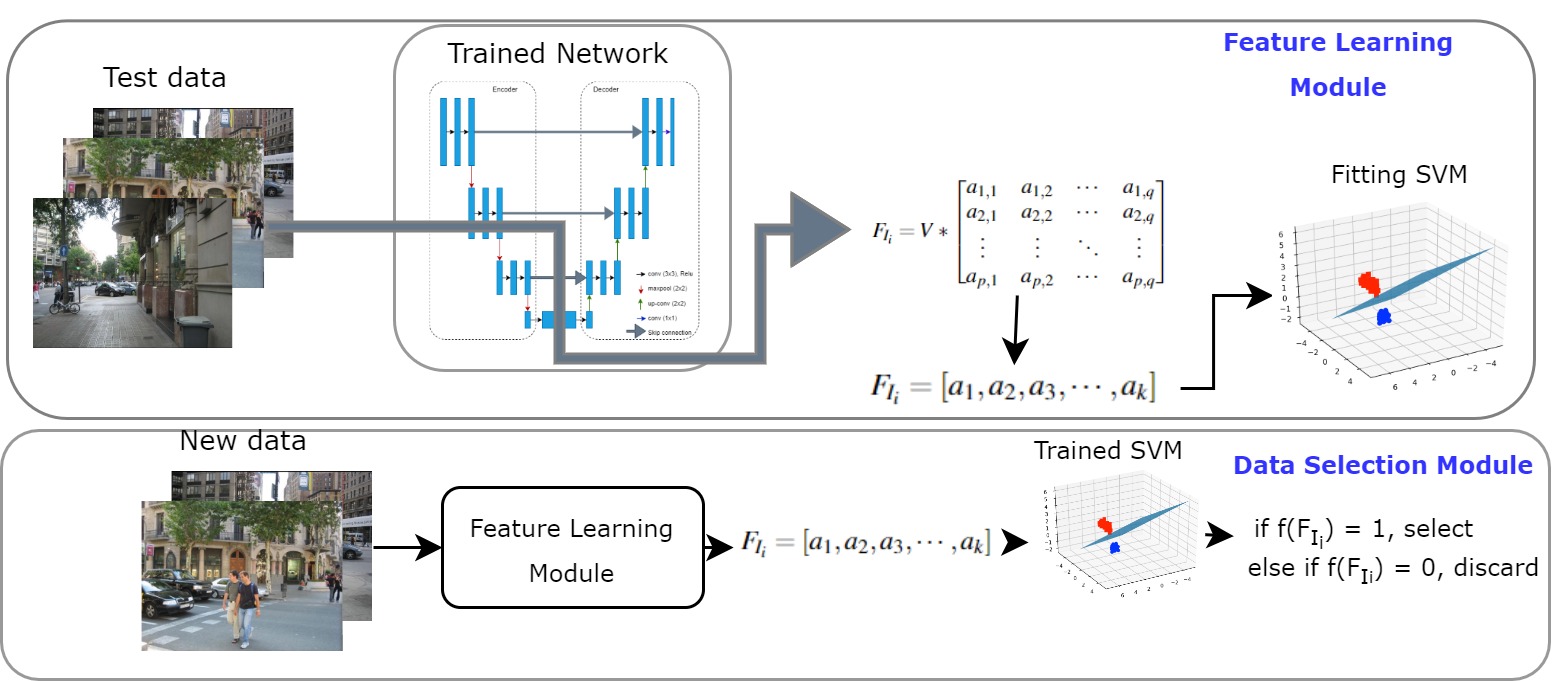}
    \caption{Illustration of the proposed approach. In the feature learning module, the feature vector representation are first learnt using an SVM on the prior segmentation network in production.  Then any new data that needs to be added for retraining the segmentation net, passes through the feature learning module and is fit to the trained SVM to predict if it must be selected for further model training or discarded. }
    \label{feature_learning}
\end{figure*}

\paragraph{Blind/Referenceless Image Spatial Quality Evaluator (BRISQUE):} In this work, we investigated BRISQUE \cite{mittal2012no}, a no-reference IQA algorithm designed to evaluate the quality of digital images without requiring a reference image for comparison. It operates by analyzing statistical properties of the image such as luminance, texture, compression etc. It spans from the idea that the distribution of pixel intensities of natural images differs from that of distorted images. First, a Mean Subtracted Contrast Normalization (MSCN) is performed as follows:

\begin{equation}
\hat{I}(i, j) = \frac{I(i, j) - \mu(i, j)}{\sigma(i, j) + C},
\end{equation}
where $i = 1, 2, ..., M$ and $j = 1, 2, ..., N$ are the spatial indices, $M$ and $N$ are the image height and width respectively, $\hat{I}$ is the resulting MSCN image. $\mu(i, j)$ is local mean field which is the Gaussian Blur of the original image. $\sigma(i, j)$ is the local variance field which is the squared difference of original image and $\mu$. MSCN normalization is effective for pixel intensities, but considering pixel relationships is crucial. So, pair-wise products of MSCN image with a shifted version of the MSCN image along four orientations: Horizontal (H), Vertical (V), Left-Diagonal (D1), Right-Diagonal (D2). The resulting five images are fitted to a Generalized Gaussian Distribution (GGD) to create a feature vector. These statistical features serve as input for a pretrained regression model, trained on a large dataset annotated by human subjects, to predict the BRISQUE score.

% is computed as follows

% \begin{equation}
% H(i, j) = \hat{I}(i, j) \hat{I}(i, j + 1)
% \end{equation}

% \begin{equation}
% V(i, j) = \hat{I}(i, j) \hat{I}(i+1, j)
% \end{equation}

% \begin{equation}
% D1(i, j) = \hat{I}(i, j) \hat{I}(i+1, j + 1)
% \end{equation}

% \begin{equation}
% D2(i, j) = \hat{I}(i, j) \hat{I}(i+1, j - 1)
% \end{equation}
% for $i = 1, 2, ..., M$ and $j = 1, 2, ..., N$. 

% The template is composed by a set of 7 files, in the
% following 2 groups:\\
% \noindent {\bf Group 1.} To format your paper you will need to copy
% into your working directory, but NOT edit, the following 4 files:
% \begin{verbatim}
%   - apalike.bst
%   - apalike.sty
%   - article.cls
%   - scitepress.sty
% \end{verbatim}

% \noindent {\bf Group 2.} Additionally, you may wish to copy and edit
% the following 3 example files:
% \begin{verbatim}
%   - example.bib
%   - example.tex
%   - scitepress.eps
% \end{verbatim}

% This section must be in one column.

\subsection{Feature vector learning based approach}

In order to ensure the reliability and improvement of the training data as we gather more data for the models, we investigate another approach, depicted in Figure \ref{feature_learning}, where we incorporate a conditioning mechanism that takes into account the data on which the current model was trained. We make use of feature vectors having the richest information extracted from the bottleneck layers of the segmentation network to train a simple Support Vector Machine (SVM) that learns to distinguish the true and false predictions made by the trained network on the test data. The purpose of this network is to guide the feature learning of the retrained models on newly acquired dataset.

Consider that we have a baseline model $M_a$ trained by an initial dataset which is assumed to be representative of our domain of interest. Then we test $M_a$ on our evaluation dataset $D_{test}$ to obtain the prediction masks.  Dice metric is computed to measure the accuracy of the obtained masks.  For each $I_i \in D_{test}$, the dice value is evaluated against a threshold to categorize each result as true or false. After passing $D_{test}$ through the trained model $M_a$ we also take the output from the bottleneck layer of the U-net and obtain a feature vector set for image $I_i \in D_{test}$, given as:

\begin{equation}
F_{I_i} = V * \begin{bmatrix}
a_{1,1}&a_{1,2}&\cdots &a_{1,q} \\
a_{2,1}&a_{2,2}&\cdots &a_{2,q} \\
\vdots & \vdots & \ddots & \vdots\\
a_{p,1}&a_{p,2}&\cdots &a_{p,q}
\end{bmatrix}
% \substack{16 \times 16}
\end{equation}
After obtaining $V$ vectors of size $p \times q$ we apply spatial pyramid pooling \cite{he2015spatial} to partition and pool information from different regions of the feature maps, and this gives an aggregated representation that retains vital contextual and positional information while reducing the dimensionality to a 1D vector for every $I_i \in D_{test}$. 

\begin{equation}
F_{I_i} = [a_1, a_2,  a_3, \cdots, a_{k}]
\end{equation}

Once we obtain the 1D vector, we use it as features to train an SVM model that learns to distinguish the vector representations for true and false predictions. 
% The labels for these vectors were assigned by considering the dice value obtained for $I_i$ upon evaluating it on $M_a$.

Subsequently, new data is then passed through the SVM, allowing for the data selection of true predictions that are then chosen for further model training. By conditioning the selection of future data on this network, we theorize that we are fine tuning the new data on the current model. Hence, we ensure that the prior knowledge of the model is preserved leading to retaining the model performance.

\section{\uppercase{Experiments}}
\begin{table*}[t]
    \centering
    \caption{Comparing the performance of the baseline model trained with $D_{orignal}$ which consists of all the initial undistorted images from Cityscapes, ADE, and VOC to the performance of the model trained with  70\% of images from $D_{orignal}$ having low BRISQUE and $30\%$ from $D_{distorted}$ having high BRISQUE score. The model performance drops when high BRISQUE scored images are added for training}
    \begin{tabular}{c|c|ccc} \hline 
         No.&  Training Dataset&  Dice&  PR score&  F-score\\ \hline \hline
         1&  $D_{orignal}$ (Baseline) & 0.404864 & 0.589132 & 0.289338\\ 
         2 & 2/3 $D_{orignal}$ + 1/3 $D_{distorted}$ & 0.369254 & 0.549964 & 0.36924 \\ \hline 
    \end{tabular}
    \label{tab:distorted_exp_sel}
\end{table*}

\begin{table*}
    \centering
    \caption{Comparing the performance of the model trained with $D_{distorted}$ which consists of distorted versions of all images from Cityscapes, ADE, and VOC to the performance of the model trained with $D'_{distorted}$ which contains selected images having BRISQUE $<$ $60$, meaning higher quality images. The model performance better when we prune images having high BRISQUE}
    \begin{tabular}{c|c|c c c} \hline 
         No.&  Training Dataset&  Dice&  PR score&  F-score\\ \hline \hline
         1&  $D_{distorted}$ & 0.309860 & 0.386605 & 0.208074\\ 
         2 & $D'_{distorted}$ & \textbf{0.371031} & \textbf{0.484460} & \textbf{0.26014}5\\ \hline 
    \end{tabular}
    \label{tab:distorted_exp}
\end{table*}
\subsection{Datasets and evaluation methods}

\paragraph{Datasets used:} We use three benchmark datasets for semantic segmentation namely the Adverse Environment Conditions dataset (ADE20K) \cite{zhou2017scene} \cite{zhou2019semantic}, Cityscapes \cite{cordts2016cityscapes}, and PASCAL Visual Object Classes (VOC) \cite{Everingham10} dataset. ADE20K consists of $480 \times 600$ images belonging to nearly $150$ classes. Cityscapes contains annotations for $30$ classes and VOC comprises $20$ classes.  For the sake of testing our approach, we focus on segmenting one object of interest from these datasets. Before passing the images to train the model, they are resized to $256 \times 256$ with $3$ channels. In total the training data comprises of $800$ images from ADE20K, $800$ from Cityscapes and 300 from VOC. The test dataset $D_{test}$ is created by taking $100$ images from each of these $3$ datasets that are not part of the training data. The initial results obtained on $D_{test}$ using the U-net model trained with different combinations of the $3$ datasets as training sets is given in Table \ref{tab:dataset_metrics}. 

\paragraph{Evaluation metrics:} To evaluate the models we employ three main metrics: the dice coefficient, the Area Under the Curve (AUC) of Precision-Recall (PR score), and the F-score. The dice coefficient quantifies the degree of overlap between our model's predicted binary segmentation mask and the ground truth, providing a measure of segmentation quality. Meanwhile, the AUC of the PR curve offers a comprehensive assessment of binary classification performance, capturing the trade-off between precision and recall at various thresholds. Lastly, the F-score, which is the harmonic mean of precision and recall, provides a balanced evaluation, particularly valuable when dealing with class imbalances. 

\subsection{Model architecture}

% Distortion logic  

% \begin{figure}
% \begin{centering}
% \includegraphics[width=\columnwidth]{images/unet.png}
% \par\end{centering}
% \caption{U-net model architecture: A schematic representation of the U-Net model, a convolutional network with a U-shaped encoder-decoder structure, featuring skip connections for segmentation tasks.
% }
% \label{unet}
% \end{figure}

The semantic segmentation model architecture on which all the experiments are performed is the U-net \cite{ronneberger2015u}. The U-net follows an encoder-decoder architecture, wherein the encoder extracts high-level features through a contracting path involving operations like convolution and max-pooling. The decoder, mirroring the encoder, employs transposed convolutions to gradually increase spatial dimensions. Skip connections to corresponding encoder layers aid in preserving details. During training the binary cross-entropy loss function is minimized.

% During training, it minimizes the binary cross-entropy loss function, given as:

% \begin{equation}
%     L(y, \hat{y}) = - \left( y \cdot \log(\hat{y}) + (1 - y) \cdot \log(1 - \hat{y}) \right),
% \end{equation}
% where $y$ is the true label (either 0 or 1) and $\hat{y}$ is the predicted probability that the input belongs to class 1.

% eplain teh optimizing parameters

\section{\uppercase{Results and Discussion}}

\subsection{Approach 1: Results on IQA metrics based selection}

\begin{figure}[t]
\begin{centering}
\includegraphics[width=\columnwidth, height=2.5in]{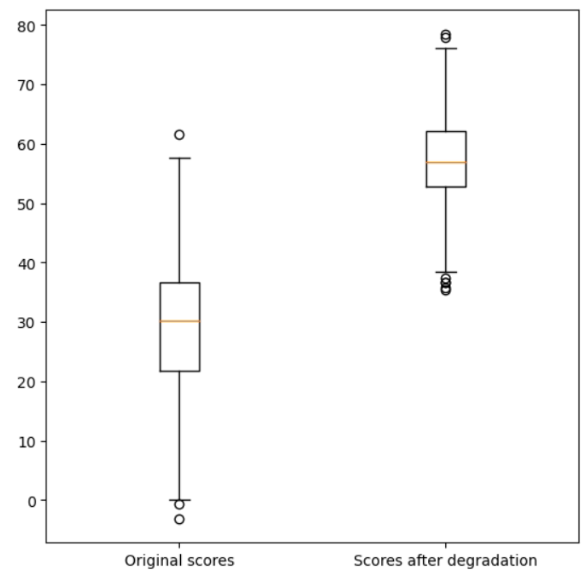}
\par\end{centering}
\caption{Distribution of original images, $D_{original}$ BRISQUE scores and BRISQUE scores for images after image degradation ($D_{distorted}$). Lower BRISQUE score indicate better quality. The original images distribution reflects an overall high perceptual quality images. After degradation, we see an overall increase in BRISQUE due to a poor quality.
}
\label{original_brisque}
\end{figure}

To test the IQA metric based selection, we first need to assess where our data stands in terms of quantitative quality. Hence we compute the BRISQUE, for all the images in the three datasets as shown in Figure \ref{original_brisque}. This distribution shows that the perceptual quality of these images is good. However, this is not always the case in real world scenarios where we might end up acquiring severely degraded data and retraining the model using this noisy data. Hence, to assess if using IQA metrics aids in filtering out the noisy data, we first distort theses images to degrade their quality. Consider our original data pool denoted by $D_{original}$, with every image of dimension $M \times N$, i.e. $I_{M\times N}\in D_{original}$, we perform a pyramidal downscaling followed by a pyramidal upscaling. For each pixel $(i, j)$ in the downscaled image $I'$, the corresponding pixel value is computed by averaging the values of the four neighboring pixels in $I$ as follows:

\begin{align*}
I'(i,j) & =\frac{I(i-1,j+1)+I(i+1,j+1)}{4} \tag{11}\\
 & \,\,\,\,+\frac{I(i-1,j-1)+I(i+1,j-1)}{4},
\end{align*}
where $i = 1,...,M/2$ and $j=1,...,N/2$. Then, in upscaling, for each pixel $(i, j)$ in the desired upscaled image $I''$ (where $i$ ranges from $1$ to $M$ and $j$ ranges from $1$ to $N$), the corresponding pixel value is directly copied from the nearest pixel in the downscaled image $I'$ as follows:

\begin{equation*}
    I''(i, j) = I'( (i/2), (j/2) ), i \in M, j \in N \tag{12}
\end{equation*}

The final distorted image $I''$ will have a blurring effect. The IQA metric values for the distorted set $D_{distorted}$ are depicted in Figure \ref{original_brisque}, where we can see a clear degradation in the values reflecting this reduction in quality. 

% \begin{figure}[t]
% \begin{centering}
% \includegraphics[width=0.6\columnwidth]{images/brisque_scores_distorted.png}
% \par\end{centering}
% \caption{Distribution of BRISQUE scores for images from $D_{distorted}$ after image degradation. 
% }
% \label{distorted_brisque}
% \end{figure}

\begin{table*}
    \centering
    \caption{Initial performance on $D_{test}$ of the U-net model trained using different combinations of the three datasets.}
    \begin{tabular}{c|c|c c c} \hline 
         No.&  Training Dataset&  Dice&  PR score&  F-score\\ \hline \hline 
         1&  Cityscapes &   0.375370 &  0.580734& 0.307178\\ 
         2 & ADE & \textbf{0.513763} & 0.645072 & \textbf{0.391683} \\ 
3 & VOC & 0.245870 & 0.459835 & 0.154959 \\ 
        4 & Cityscapes + VOC & 0.286961 & 0.545884 & 0.186165 \\ 
5 & Cityscapes + ADE & 0.493080 & \textbf{0.645643} & 0.370273 \\ 
6 & ADE + VOC & 0.389131 & 0.617254 & 0.276805 \\
7 & Cityscapes + VOC + ADE & 0.404864 & 0.589132 & 0.289338 \\ \hline 
    \end{tabular}
    \label{tab:dataset_metrics}
\end{table*}

\begin{table*}[t]
    \centering
    \caption{Performance of trained U-nets on $D_{test}$ using the feature vector learning approach. The first row shows the initial performance of the model on ADE dataset. As more data is added to the pipeline without any data selection strategy, the model performance degrades as seen in row 2, due to noisy data. Upon applying the selection strategy on the new datasets, we see that it preserves the prior knowledge and prevents further performance degradation in row 3.}
    \begin{tabular}{c|c|c c c} \hline 
         No.&  Training Dataset&  Dice&  PR score&  F-score\\ \hline \hline 
        1 & ADE & \textbf{0.513763} & \textbf{0.645072} & \textbf{0.391683} \\ 
        2 & ADE + Cityscapes + VOC (Baseline Model) & 0.404864 & 0.589132 & 0.289338 \\ 
        3 & ADE + $(Cityscapes + VOC)_{Selected Data}$  & \textbf{0.462565} & \textbf{0.618605} & \textbf{0.342391} \\ \hline 
    \end{tabular}
    \label{tab:feature_based_results}
\end{table*}

To investigate how adding high valued BRISQUE scored images to the dataset pipeline affect the model performance, we perform an experiment where we include $70\%$ images from $D_{orignal}$, and $30\%$ of images from $D_{distorted}$. The performance achieved by conducting this experiment is demonstrated in Table \ref{tab:distorted_exp_sel}. It is observed that when we have images having high BRISQUE scores (meaning poor quantitative quality), the model performance degrades as we see the PR score drops to $0.549964$ from $0.589132$ as compared to the baseline model trained purely with all $D_{orignal}$. This indicates a decreased ability to correctly classify true positives while maintaining a similar rate of false positives, due to the introduction of lower-quality images.

% Further investigations are conducted to study the impact of data quality on the model training process. In this analysis, we focus on datasets comprising distorted images, specifically $D_{distorted}$ and $D'_{distorted}$. The former encompassed distorted versions of all images from Cityscapes, ADE, and VOC, while the latter is carefully curated to include only selected images with BRISQUE scores below $60$, indicative of higher quality. The model's performance, as summarized in Table \ref{tab:distorted_exp}, exhibits a noteworthy contrast between the two datasets. When trained with $D_{distorted}$, the model achieves a dice coefficient of $0.309860$, a PR score of $0.386605$, and an F-score of $0.208074$. Conversely, utilizing $D'_{distorted}$ for training leads to significantly improved results, with a dice coefficient of $0.371031$, a PR score of $0.484460$, and an F-score of $0.260145$. Hence, the model's performance exhibits improvement when images with high BRISQUE scores were pruned, indicating the substantial benefit of data curation using IQA metrics to ensure high-quality inputs for model training. 

Further investigations explore the impact of data quality on model training, focusing on distorted images in datasets $D_{distorted}$ and $D'_{distorted}$. The former includes distorted versions of Cityscapes, ADE, and VOC images, while the latter selectively includes high-quality images with BRISQUE scores below $60$. Table \ref{tab:distorted_exp} summarizes the model's performance on these datasets. Training with $D_{distorted}$ yields a dice coefficient of $0.309860$, a PR score of $0.386605$, and an F-score of $0.208074$. Contrastingly, utilizing $D'_{distorted}$ results in significantly improved metrics: a dice coefficient of $0.371031$, a PR score of $0.484460$, and an F-score of $0.260145$. This improvement underscores the substantial benefit of data curation using IQA metrics, ensuring high-quality inputs for model training.

\subsection{Approach 2: Results on Feature Learning based method}

% \begin{table}
%     \centering
%     \caption{Dataset size statistics.}
%     \begin{tabular}{c|c} \hline 
%     Data &  Number of Images  \\ \hline \hline
%         ADE&  800 \\ 
%         Cityscapes + VOC & 1100 \\ 
%         Selected from Cityscapes + VOC & 550 \\ \hline 
%     \end{tabular}
%     \label{tab:data_statistics}
% \end{table}

\begin{table*}
    \centering
    \caption{Comparing results of the two approaches: $(Cityscapes + VOC)_{F}$ uses feature learning, and $(Cityscapes + VOC)_{Q}$ involves quality-based selection with BRISQUE $T<40$. Both are compared to the baseline model.}
    \begin{tabular}{c|c|c c c} \hline 
         No.&  Training Dataset&  Dice&  PR score&  F-score\\ \hline \hline 
         1&  ADE+ Cityscapes + VOC (baseline) & 0.404864 & 0.589132 & 0.289338\\ 
         2 & ADE+ $(Cityscapes + VOC)_{F}$  & \textbf{0.462565} & 0.618605 & \textbf{0.342391} \\ 
3 & ADE+ $(Cityscapes + VOC)_{Q}$ & 0.437014 & \textbf{0.626327} & 0.322737 \\ \hline 
    \end{tabular}
    \label{tab:comapring_brisque_svm}
\end{table*}

In this approach, we first perform a systematic exploration of various training datasets for U-Net models and examine their performance on our test dataset $D_{test}$. Our initial objective is to identify a benchmark model for our proposed method. To this end, we trained and evaluated multiple U-Net models using different combinations of three diverse datasets, ADE, Cityscapes, and VOC. The performance results of these models are given in Table \ref{tab:dataset_metrics}. Among these, we observed that the U-Net model trained solely on the ADE dataset exhibits the best performance, achieving a dice coefficient of $0.513763$, PR score of $0.645072$, and and F-score of $0.391683$. Hence, we consider this model as our preferred choice for the initial production model in our deployment environment.

Spanning from our initial formulated problem, we add Cityscapes and VOC, to the initial dataset pipeline, which was trained using only ADE. It can be seen that the model suffers degradation in performance, shown by the dice value falling from from $0.513763$ to $0.404864$ and the PR score dropping from $0.645072$ to $0.589132$. To handle this drop, we introduce our feature vector learning approach, which involves leveraging the ADE-trained U-Net as a base for selecting images from additional datasets, Cityscapes and VOC, to be added to the retraining of the model. 

\begin{figure}[!h]
\begin{centering}
\includegraphics[width=\columnwidth]{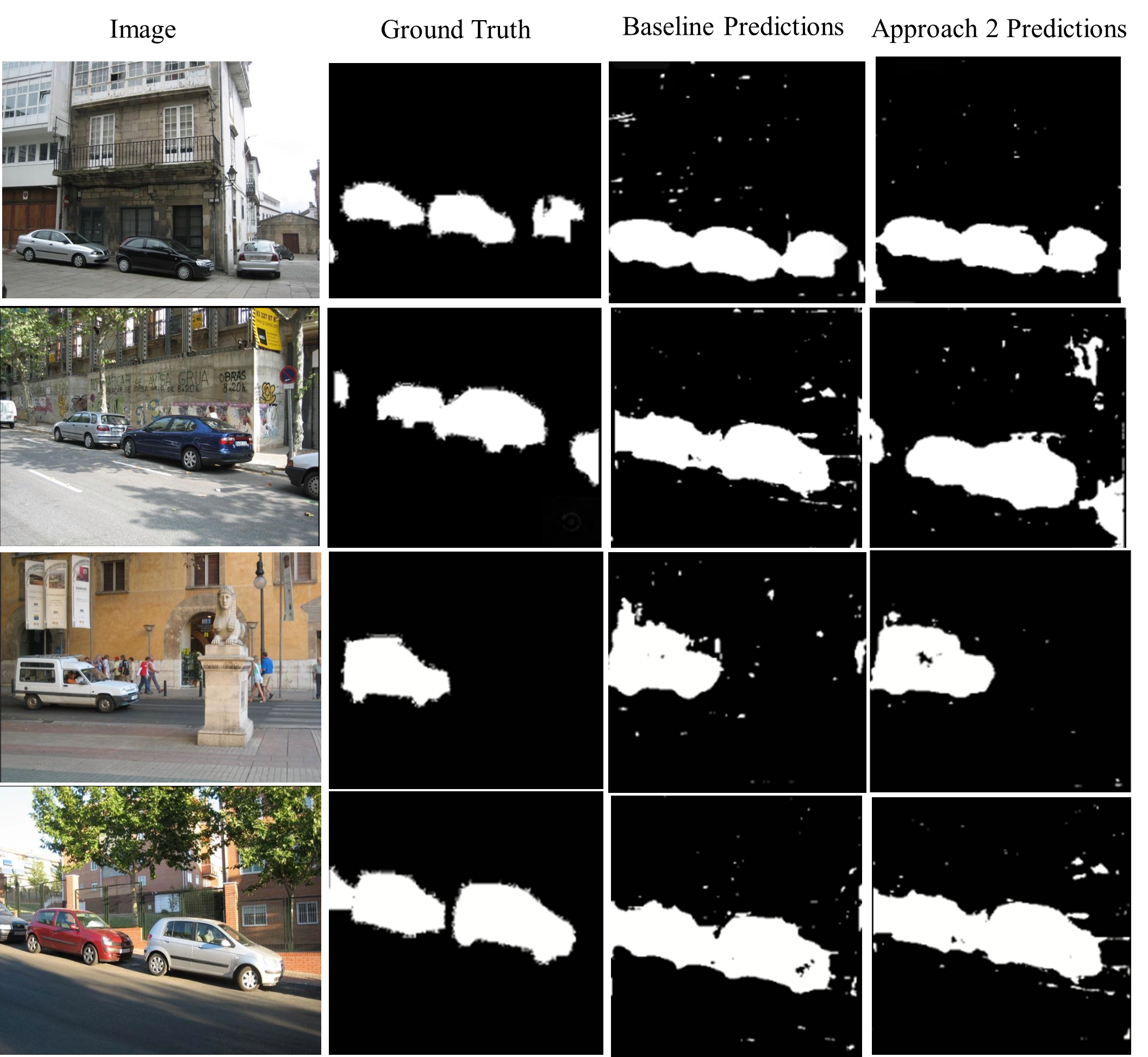}
\par\end{centering}
\caption{Visualizing predictions made by the trained models. Approach 2 predictions are shown in the last column of the image, where we see a visual reduction of the false positives compared with the baseline model predictions.
}
\label{feature_based_results}
\end{figure}

The results of this selection are illustrated in Table \ref{tab:feature_based_results}. Table \ref{tab:data_statistics} presents the number of images selected by this approach. Applying the feature-based data selection strategy to new datasets yields substantial performance improvements, evident in higher dice coefficient, PR, and F-score metrics. This improvement surpasses the baseline model (ADE + Cityscapes + VOC), which uses all available images without any selection. Our method not only maintains high performance with the ADE dataset but also adapts effectively to diverse data sources. This adaptability is crucial in real-world scenarios with continuous dataset updates, showcasing the robustness and practical applicability of our proposed methodology for semantic segmentation tasks. Some prediction visualizations are depicted in Figure \ref{feature_based_results}. 

% These findings underscore the efficacy of our approach in enhancing the model's ability to generalize and segment objects accurately.

\subsection{Comparing the two approaches}
\begin{table}
    \centering
    \caption{Dataset size statistics.}
    \begin{tabular}{c|c} \hline 
    Data &  \# of Images  \\ \hline \hline
        ADE&  800 \\ 
        Cityscapes + VOC & 1100 \\ 
        Selected from Cityscapes + VOC & 550 \\ \hline 
    \end{tabular}
    \label{tab:data_statistics}
\end{table}
Both approaches demonstrate an overall performance improvement, as shown in Table \ref{tab:comapring_brisque_svm}. The quality-based method excels in reducing false positives, as indicated by slight improvements in PR scores and F-scores. On the other hand, the feature vector learning-based approach enhances dice, F-score, and reduces false predictions, though not as prominently as the quality-based method. If the primary goal is to minimize false predictions, the quality-based approach appears to be the preferred choice, showcasing its efficacy.

\section{\uppercase{Conclusions}}
\label{sec:conclusion}

In this work we investigate two strategies to handle model drift: data quality assessment and data conditioning based on prior model knowledge. The former relies on image quality metrics to meticulously select high-quality training data, thereby bolstering model robustness. In contrast, the latter leverages learned feature vectors from existing models to guide the selection of future data, aligning it with the model's prior knowledge. Through extensive experimentation, we provide valuable insights into the effectiveness of these approaches in enhancing the performance and reliability of semantic segmentation models. These findings underscore the significance of data quality and alignment with prior knowledge in sustaining the efficacy of AI models in dynamic real-world environments, thus contributing to the ongoing advancement of computer vision capabilities. Person Re-Identification, a critical computer vision task, has also seen severe model drift under incremental scenario \cite{Khaldi_2024_WACV,Nguyen_2024_SEMI,Nguyen_2024_ASGL}. In future works, we intend to leverage our approach to propose methods to tackle this problem.

\bibliographystyle{apalike}
{\small
\bibliography{main}}

% \section*{\uppercase{Appendix}}

% If any, the appendix should appear directly after the
% references without numbering, and not on a new page. To do so please use the following command:
% \textit{$\backslash$section*\{APPENDIX\}}

\end{document}